%% file: main.tex
\documentclass[conference,letterpaper]{IEEEtran}
\usepackage{pifont}    

\newcommand{\equalcontrib}{\textsuperscript{*}}
\newcommand{\corr}{\textsuperscript{\dag}}  
\newcommand{\UToronto}{\textsuperscript{\ddag}}
\newcommand{\UWaterloo}{\textsuperscript{\S}}
\usepackage[utf8]{inputenc} 
\usepackage[T1]{fontenc}
\usepackage{graphicx}
\usepackage{url}
\usepackage{ifthen}
\usepackage{cite}
\usepackage[cmex10]{amsmath} 
\usepackage{cleveref}
\usepackage{booktabs}
\usepackage{multirow}
\usepackage{amsmath,amssymb}
\usepackage{mathrsfs} 
\usepackage{xcolor}
\usepackage{color, colortbl}
\usepackage{mwe}
\usepackage{wrapfig} 
\definecolor{cmtgray}{gray}{0.55}
\definecolor{mygray}{gray}{.8}
\definecolor{lightgray}{gray}{0.93}
\definecolor{Mygray}{gray}{0.35}

\usepackage[normalem]{ulem}
\usepackage{algorithm}
\usepackage{algpseudocode}
\usepackage{dsfont}
\algrenewcommand\textproc{\textsc} 
\algrenewcommand\algorithmicrequire{\textsc{Input:}}
\algrenewcommand\algorithmicensure{\textsc{Output:}}
\algrenewcommand{\algorithmiccomment}[1]{\hfill\textcolor{cmtgray}{\#~#1}}

\newtheorem{theorem}{Theorem}
\makeatletter
\renewcommand{\ALG@name}{Algorithm}
\makeatother

\begin{document}

\title{Normalized Conditional Mutual Information Surrogate Loss for Deep Neural Classifiers} 

\author{\IEEEauthorblockN{Linfeng Ye\UWaterloo\UToronto\equalcontrib,
Zhixiang Chi \UToronto\equalcontrib,
Konstantinos N. Plataniotis\UToronto,
En-hui Yang\UWaterloo\corr}
\IEEEauthorblockA{\UWaterloo Department of Electrical and Computer Engineering, University of Waterloo, Waterloo, Canada}
\IEEEauthorblockA{\UToronto The Edward S. Rogers Sr. Department of Electrical and Computer Engineering,
University of Toronto, Toronto, Canada
}
Email: \{l44ye, ehyang\}@uwaterloo.ca\UWaterloo,
\{chizhixi, kostas\}@ece.utoronto.ca\UToronto
\thanks{\equalcontrib~Equal contribution.}
\thanks{\corr~Corresponding author.}
}
\IEEEoverridecommandlockouts
\maketitle

\begin{abstract}
 
 In this paper, we propose a novel information-theoretic surrogate loss—normalized conditional mutual information (NCMI)—as a drop-in alternative to the de facto cross-entropy (CE) for training deep neural network (DNN)-based classifiers. We first observe that the model’s NCMI is inversely proportional to its accuracy. Building on this insight, we advocate to use NCMI as the surrogate loss for DNN classifier, and propose an alternating algorithm to efficiently minimize the NCMI. Across natural image recognition and whole-slide imaging (WSI) subtyping benchmarks, NCMI-trained models surpass state-of-the-art losses by substantial margins at a computational cost comparable to that of CE. Notably, on ImageNet, NCMI yields a 2.77\% top-1 accuracy improvement with ResNet-50 comparing to the CE; on CAMELYON-17, replacing CE with NCMI improves the macro-F1 by 8.6\% over the strongest baseline. Gains are consistent across various architectures and batch sizes, suggesting that NCMI is a practical and competitive alternative to CE. All code and data are publicly available at \url{https://github.com/Linfeng-Ye/NCMI}.
\end{abstract}


\section{Introduction}

Deep neural classifiers are typically trained by mapping an input $\boldsymbol{x}$ to a feature vector $\boldsymbol{z}=f_\theta(\boldsymbol{x})\in\mathbb{R}^m$, followed by a linear classifier $h$ whose softmax produces a distribution on the \emph{class} simplex. In this de facto setting, cross-entropy (CE) is used as a principled surrogate loss which optimizes class probabilities in a $c$-dimensional probability space determined by the logits.

A large body of work augments CE with additional regularizers \cite{ranasinghe2021orthogonal, hui2023cut, CMI} or proposes ad-hoc alternatives \cite{hui2021evaluation, khosla2020supervised} that empirically compete with CE.
However, these approaches typically remain CE-centric; their performance depends on CE, while the auxiliary terms are ineffective on their own,  require extensive hyperparameter tuning, or rely on very large batch sizes to realize their gains.

Furthermore, much of the CE literature implicitly operates in an overparameterized regime in which the penultimate feature dimension $m$ is comparable to (or larger than) the number of classes $c$. When $m$ is constrained---e.g., due to compact embeddings for memory/latency or architectures that intentionally keep $m$ small---the terminal linear map induces a low-rank family of class-score vectors whose rank is at most $m$, which becomes a geometric limitation when $c \gg m$ \cite{NEURIPS2019_78f7d96e}. Closely related expressiveness limits of dot-product--Softmax parameterizations have been formalized as the \emph{Softmax bottleneck}: if the intrinsic rank of the target log-probability matrix exceeds the embedding dimension, no choice of parameters can represent the true conditional distributions \cite{yang2018breaking}. In such constrained-dimension regimes, the usual “CE-on-class-probabilities” viewpoint provides limited guidance on how to shape the representation itself, and the separation intuitions that hold in high-dimensional settings no longer directly apply.


In this paper, we apply a feature-induced simplex formulation. We view a DNN together with a normalized sigmoid function (NSF) as $(\sigma^{NSF}\circ f_\theta): \boldsymbol{x}\to \boldsymbol{p}\in\mathcal{P}^m$, and inference can be performed via centroid comparison in KL divergence. Our goal is to learn simplex outputs that are (i) highly concentrated within each class and (ii) well separated across classes. Following the information-theoretic modeling in \cite{CMI}, we represent classification as a Markov chain and quantify the concentration of the output distribution using the conditional mutual information (CMI) between the input and output, conditioned on the ground-truth label, and we quantify separation with $\Gamma$ (see \Cref{Sec:MarkovChain}) and define the normalized conditional mutual information (NCMI) as the ratio between the CMI and $\Gamma$. Compared with \cite{CMI}, where NCMI enters the learning problem as a constraint, we train the DNN classifier directly by minimizing NCMI.

Beyond conceptual alignment, NCMI is practically effective: across pretrained ResNet variants, NCMI exhibits a near-linear inverse relationship with top-1 accuracy, suggesting it is an informative performance surrogate. We further develop an efficient alternating minimization procedure for NCMI optimization. Finally, because mapping features to a simplex can suffer from \emph{single-mode collapse} (e.g., softmax-induced degeneracy), we incorporate feature centering and a NSF to stabilize training and prevent collapse.

\textbf{Contributions:} (i) an information-theoretic surrogate loss (NCMI) tailored to DNN-based classifiers; (ii) an efficient alternating algorithm for minimizing NCMI; and (iii) comprehensive validation on natural-image and whole-slide imaging benchmarks.

\section{Related work}
Within the existing literature, CE and its variants are the de facto objectives for classification. Several works have attempted to improve CE. Empirical studies report that DNNs with compact feature clusters usually outperform those with sparse clusters \cite{oyallon2017building, JMLR:v21:20-933, papyan2020prevalence}. These insights have been further analyzed under the Gaussian-mixture assumption on the feature distribution \cite{zarka2021separation}. Building on this line of work, subsequent works augment CE with regularizers. Specifically, Hui et al. \cite{hui2023cut} add an $\ell_2$ penalty to the non-ground-truth entries of the predicted probability distribution, and OPL \cite{OPL} explicitly clusters same-class features while enforcing orthogonality between different classes in the penultimate layer.

Another line of work improves classification accuracy by modifying CE. Focal Loss \cite{lin2017focal} down-weights well-classified samples via a power transformation so training emphasizes hard instances. PolyLoss \cite{leng2022polyloss} reframes standard classification losses as polynomial expansions. Hui et al. propose SquareLoss \cite{hui2021evaluation}, and empirically found that squared loss performs on par with or even outperforms CE on modern DNNs. Supervised contrastive learning (SupCon) \cite{khosla2020supervised} pulls together same-class embeddings and pushes apart different-class embeddings, followed by a linear classifier trained on the frozen features. Further, to mitigate overconfident predictions, label smoothing (LS) \cite{szegedy2016rethinking} softens the one-hot targets, which can inadvertently produce compact class clusters \cite{muller2019does}. AntiClass \cite{kat2025anticlasses} replaces the one-hot target with a one-cold target to mitigate neural collapse.

In contrast, this paper studies the surrogate loss for classification through the lens of information geometry. We view a DNN as a mapping from $\boldsymbol{x}\in \mathbb{R}^d$ to $\boldsymbol{p}\in \mathcal{P}^n$. We trains the DNN by encouraging the intra-class concentration and inter-class separation of the output distribution cluster by directly minimizing NCMI.

\section{Notation and Preliminaries}
\subsection{Notation}
\begin{figure}
    \centering
\includegraphics[width=0.8\columnwidth]{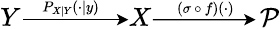}
    \caption{Mappings from the label space $Y$ to the input space $X$, and from the input space to a output space $\hat{Y}$. Input $\boldsymbol{x}$ are sampled from the class $Y=y$ according to the $P_{X|Y}(\cdot|y)$. This is further mapped by a DNN and a simplex-valued function to an output probability distribution $\boldsymbol{p}\in\mathcal{P}$.}
    \label{fig:Markov}
\end{figure}
Scalars are denoted by non-bold letters (\textit{e.g.} $\beta$), vectors by bold lowercase letter (\textit{e.g.} $\boldsymbol{a}$), the $i$-th entry of a vector $\boldsymbol{a}$ is written as $\boldsymbol{a}[i]$. We denote $\mathcal{P}^m$ as the set of all $m$-dimensional probability distributions. For any two probability distributions $\boldsymbol{p}, \boldsymbol{q} \in \mathcal{P}^m$, the Kullback–Leibler (KL) divergence is defined as 
\begin{align}
    D(\boldsymbol{p}\|\boldsymbol{q}) = \sum_{i=1}^m\boldsymbol{p}[i] \ln \frac{\boldsymbol{p}[i]}{\boldsymbol{q}[i]},
\end{align}
where $\ln$ denotes the logarithm with base $e$, the  CE is defined as
\begin{align}
    H(\boldsymbol{p},\boldsymbol{q}) = \sum_{i=1}^m\boldsymbol{p}[i] \ln \boldsymbol{q}[i],
\end{align}
write the CE of the one-hot probability distribution corresponding to $y$ and $\boldsymbol{q}$ as $H(y,\boldsymbol{q}) = -\ln \boldsymbol{q}_y$.
We denote $\mathds{1}$ as the indicator function, and define normalized sigmoid function (NSF) $\sigma^{NSF}$ and softmax function $\sigma^{SM}$ as
\begin{align}
    \sigma^{NSF}(\boldsymbol{z})[j] &= \frac{\phi(\boldsymbol{z}[j]}{\sum_{i=1}^m \phi(\boldsymbol{z}[i])},~~\phi(\boldsymbol{z}[i]) = \frac{1}{1+e^{-\boldsymbol{z}[i]}};\\
    \sigma^{SM}(\boldsymbol{z})[j] & = \frac{e^{\boldsymbol{z}[j]}}{\sum_{i=1}^m e^{\boldsymbol{z}[i]}}, \text{ where } \boldsymbol{z} \in \mathbb{R}^m.
\end{align}
Given a multi-class classification dataset $\mathcal{D}$, let $\mathcal{D}^y \subseteq \mathcal{D}$ denote the subset of samples labeled $y$, and $\boldsymbol{c_x}$ denote the label of sample $\boldsymbol{x}$.
\subsection{Modeling Classification as a Markov Chain}
\label{Sec:MarkovChain}
In a classification task with $c$ classes, a DNN $f$, a linear classifier $h$ and a $\sigma$ could be regarded as a mapping $(\sigma \circ h \circ f): \boldsymbol{x} \to \boldsymbol{p_x}$, where $\boldsymbol{x}$ is an input, and $\boldsymbol{p_x} \in \mathcal{P}^n$ is the output probability distribution. Usually, $n=c$ when we use CE as the surrogate loss. Following \cite{10619241}, we can model the classification task as a three-state Markov chain, as depicted in \Cref{fig:Markov}. As shown in \cite{CMI}, we empirically quantify the concentration of DNN's output by 
\begin{align}
    I(X; \mathcal{P}| Y) &= \frac{1}{|\mathcal{D}|}\sum_{y}\sum_{\boldsymbol{x}\in \mathcal{D}^y} D(\boldsymbol{p_x}\|\boldsymbol{s}^y),\label{Eq:CMI}\\ 
    \text{where } \boldsymbol{s}^y & \triangleq \frac{1}{|\mathcal{D}^y|} \sum_{\boldsymbol{x}\in \mathcal{D}^y} \boldsymbol{p_x},~ \text{for } y\in Y. \label{Eq:centroid}
\end{align}
\begin{figure}
    \centering
\includegraphics[width=0.8\columnwidth]{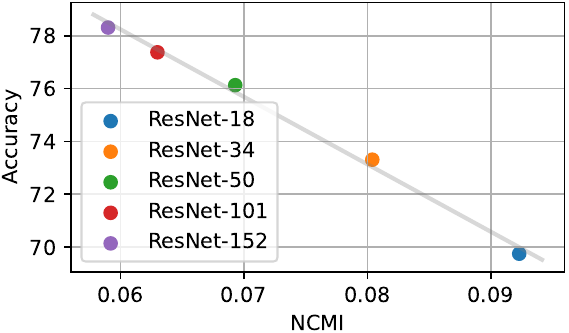}
    \caption{The accuracy vs NCMI value over the validation set of pre-trained ResNet models on the ImageNet dataset.}
    \label{fig:ACC_NCMI}
\end{figure}
Further, the separation of DNN's output can be quantified as 
\begin{align}
    \Gamma = \frac{1}{|\mathcal{D}|^2}\sum_{\boldsymbol{z}\in \mathcal{D}} \sum_{\boldsymbol{x}\in \mathcal{D}} \mathds{1}_{\{\boldsymbol{c_x}\neq \boldsymbol{c_z}\}} D(\boldsymbol{s^{\boldsymbol{c_x}}}\|\boldsymbol{p_z})\label{Eq:Gamma}
\end{align}
Ideally, we want $I(X; \mathcal{P}| Y)$ to be small while keeping $\Gamma$ large. This leads us to consider the ratio between $I(X; \mathcal{P}| Y)$ and $\Gamma$.
\begin{align}
    \hat{I}(X; \mathcal{P}| Y) \triangleq \frac{{I}(X; \mathcal{P}| Y)}{\Gamma}. \label{Eq: NCMI}
\end{align}
We refer to $\hat{I}(X; \mathcal{P}| Y) $ as the normalized conditional mutual information (NCMI). To study its relationship with classification performance, we discard the fully connected classifier of each pretrained ResNet and apply softmax to the penultimate feature vector to obtain a probability vector, from which we compute NCMI according to \Cref{Eq: NCMI}. For top-1 evaluation, we form class centroids $s^y$ as in \Cref{Eq:centroid} and classify each sample by minimum KL-divergence assignment in the induced probability space. Results on the ImageNet validation set are reported in the \Cref{fig:ACC_NCMI}. We observe a clear inverse linear relationship: models with lower NCMI achieve higher accuracy, with a Pearson correlation coefficient exceeding $-0.997$. This suggests that, for a fixed DNN family, improving performance is associated with simultaneously reducing both the error rate and the model’s NCMI during training. 
Motivated by this observation, in the next section, we demonstrate that NCMI per se suffices for training DNN classifiers.

\section{Methodology}
Previous discussions suggest a new surrogate loss for training DNN-based classifiers. Specifically, in the learning process, instead of minimizing the CE surrogate loss, which pulls the output toward predefined distributions, we aim to minimize $\hat{I}(X; \mathcal{P}| Y)$. The algorithm for minimizing the NCMI is outlined below.

\subsection{Training DNN by minimizing NCMI}
The optimization problem can be written as 
\begin{align}
    &\min_{\boldsymbol{\theta}}~ \hat{I}(X; \mathcal{P}| Y) = \nonumber \\ 
    &\min_{\boldsymbol{\theta}} \frac{\frac{1}{|\mathcal{D}|}\sum_{y} \sum_{\boldsymbol{x}\in \mathcal{D}^y} D(\boldsymbol{p_x}\|\boldsymbol{s}^y)}{\frac{1}{|\mathcal{D}|^2}\sum_{\boldsymbol{z}\in \mathcal{D}} \sum_{\boldsymbol{x}\in \mathcal{D}} \mathds{1}_{\{\boldsymbol{c_x}\neq \boldsymbol{c_z}\}} D(\boldsymbol{s^{\boldsymbol{c_x}}}\|\boldsymbol{p_z})} \label{Eq:NCMI_Object}
\end{align}
We notice that the objective in \Cref{Eq:NCMI_Object} is not amenable to parallel computation via GPU due to the dependency of $\hat{I}(X; \mathcal{P}|Y)$ on the centroid $\boldsymbol{s}^y$ of each cluster corresponding to $Y=y$ (see \Cref{Eq:centroid}). To overcome this, we introduce a dummy distribution $\boldsymbol{q}^y\in \mathcal{P}^n$ for each $y\in [C]$ and convert it into a double minimization problem, as shown in the following theorem, which will be proved in the Appendix.
\begin{theorem} 
\label{theorem1}
For any DNN: $\boldsymbol{x} \to \boldsymbol{p}$, 
\begin{align}
    &\min_{\boldsymbol{\theta}} \frac{\frac{1}{|\mathcal{D}|}\sum_{y} \sum_{\boldsymbol{x}\in \mathcal{D}^y} D(\boldsymbol{p_x}\|\boldsymbol{s}^y)}{\frac{1}{|\mathcal{D}|^2}\sum_{\boldsymbol{z}\in \mathcal{D}} \sum_{\boldsymbol{x}\in \mathcal{D}} \mathds{1}_{\{\boldsymbol{c_x}\neq \boldsymbol{c_z}\}}D(\boldsymbol{s^{\boldsymbol{c_x}}}\|\boldsymbol{p_z})}\equiv \nonumber\\ 
& \min_{\boldsymbol{q}^v, v\in [C]} \min_{\boldsymbol{\theta}} \bigg[\frac{1}{|\mathcal{D}|}\sum_{y} \sum_{\boldsymbol{x}} D(\boldsymbol{p_x}\|\boldsymbol{q}^y)\bigg]/ \nonumber\\
&\bigg[\frac{1}{|\mathcal{D}|^2}\sum_{\boldsymbol{z}\in \mathcal{D}} \sum_{\boldsymbol{x}\in \mathcal{D}} \mathds{1}_{\{\boldsymbol{c_x}\neq \boldsymbol{c_z}\}} \Big[H(\boldsymbol{p_x},\boldsymbol{p_z})-H(\boldsymbol{p_x}, \boldsymbol{q}^{\boldsymbol{c_z}})\Big]\bigg] \label{Eq:doubleMin}
\end{align}
\end{theorem}
By reformulating the single minimization as a double minimization, \Cref{Eq:doubleMin} suggests an alternating algorithm. With the centroids $\boldsymbol{q}^v, v\in [C]$ fixed, we update the DNN parameters $\theta$ by stochastic  gradient descent (SGD) on the mini-batch objective. 
Then with $\theta$ fixed, the exact minimizer for each $\boldsymbol{q}^v$ is the class centroid of the output probability vectors associated with label $v$. Computing these centroids in closed form after every $\theta$-update require dataset-wide aggregation. To keep training efficient, we instead update centroids by SGD. 
Concretely, we parameterize the centroid $\boldsymbol{q}^v, v\in [C]$ by an unconstrained vector $\boldsymbol{\xi}^v$ and map it to the simplex using softmax, $\boldsymbol{q}^v = \sigma^{NSF}(\boldsymbol{\xi}^v)$; with $\theta$ fixed, we perform one SGD step on $\boldsymbol{\xi}^v, v\in [C]$ using the same mini-batch loss.

\begin{algorithm}
\caption{PyTorch-style pseudo-code of the proposed alternating algorithm for solving the optimization problem in \Cref{Eq:doubleMin}.}
\label{Alg:NCMI}
\begin{algorithmic}[1]
\Statex \Comment{model $f_{\boldsymbol{\theta}}$; centroid $\boldsymbol{\xi}$; momentum rate $m$; temperature $\tau$; \text{centroid and model optimizer }$\text{optimizer}_{\boldsymbol{\xi}},  ~\text{optimizer}_{\boldsymbol{\theta}}$.}
\For{$\boldsymbol{x},~ \boldsymbol{y}$ in loader}
  \State $\boldsymbol{z} \gets f_{\boldsymbol{\theta}}(\boldsymbol{x})- \boldsymbol{c}$ \Comment{$\boldsymbol{z}.\text{shape: } [B, D]$}
  \State $\boldsymbol{z}' \gets \textsc{L2Normalize}(\boldsymbol{z})/\tau$   \Comment{$\ell_2$ norm / temperature}
  \State $\boldsymbol{c} \gets m\ast \boldsymbol{c} + (1-m)\ast \boldsymbol{z}.\text{mean}(\text{dim=}0).\text{detach}()$ 
  \Statex \Comment{$\boldsymbol{c}.\text{shape: } [1, D]$}
  \State $\boldsymbol{p}, \boldsymbol{q}\gets \sigma^{\text{NSF}}(\boldsymbol{z'}),\ \sigma^{\text{NSF}}(\boldsymbol{\xi})$ 
    \Statex \Comment{$\boldsymbol{p}.\text{shape: } [B, D]; ~\boldsymbol{q}.\text{shape: } [C, D]$}
  \State $\text{Calculate }\textbf{NCMI} \text{ according to \Cref{Eq:doubleMin}}$
  \State $\text{optimizer}_{\boldsymbol{\xi}}.\text{zero\_grad}(),~ \text{optimizer}_{\boldsymbol{\theta}}.\text{zero\_grad}()$
  \State $\text{\textbf{NCMI}}.\text{backward()}$
  \State $\text{optimizer}_{\boldsymbol{\xi}}.\text{step}(),~ \text{optimizer}_{\boldsymbol{\theta}}.\text{step}()$
\EndFor
\end{algorithmic}
\end{algorithm}
In the next section, we present the details of the training recipe and the evaluation protocols used to assess the NCMI loss.

\subsection{Implementation and evaluation protocols}
In this section, we provide the implementation details for training using NCMI and present the evaluation protocols applied in our experiments.

\paragraph{NCMI training} A network $f_{\boldsymbol{\theta}}$\footnote{We use ResNet \cite{He_2016_CVPR} for image recognition and multiple instance learning models \cite{ilse2018attention} for whole slide image.} maps input to a feature vector, followed by $\ell_2$ normalization, which is further mapped to a probability distribution.  To avoid the model's output probability distribution collapsing to a single distribution, following the \cite{caron2021emerging, oquab2024dinov}, we center the feature, then scale it with a pre-defined temperature $\tau$, then we use NSF to map the feature vectors to probability vectors. 
We present PyTorch-style \cite{paszke2019pytorch} pseudo-code for NCMI implementation in \Cref{Alg:NCMI}. Please refer to our code repository for full training details.
\paragraph{Evaluation Protocols}
We evaluate the performance of the NCMI-trained model under two protocols: linear probing and decision-based on comparison with centroids.

\textit{Linear probing.} We first evaluate the NCMI-trained model with the standard protocol by training a linear classifier on frozen features \cite{khosla2020supervised} using CE. We apply the same data augmentation as in the training process, freeze the model trained by NCMI, drop the $\sigma^{\text{NSF}}$, and train a linear classifier using stochastic gradient descent (SGD).

\textit{Decision based on centroids comparison.}
We further evaluate the NCMI-trained model on unseen samples from the test set by comparing them with the centroid of each class. To this end, the NCMI trained model predicts output based on comparison with centroids, specifically, we calculate the KL divergence between its output distribution $\boldsymbol{p}$ of the model and each centroid $\boldsymbol{q}^v$ per class $D(\boldsymbol{p}\|\boldsymbol{q}^v),~ v\in[C]$. Then the prediction is made based on the centroid with the smallest KL-divergence value.
\section{Experiments}
To illustrate the effectiveness of NCMI and compare it with state-of-the-art CE alternatives, a series of experiments was conducted. We conduct experiments on two widely used natural image datasets, namely CIFAR-100 \cite{krizhevsky2009learning} and ImageNet \cite{5206848}, as well as two whole-slide image datasets, namely CAMELYON-17 \cite{8447230} and BRACS \cite{brancati2022bracs}. In the tables, \texttt{NCMI-LP} and \texttt{NCMI-CC} denote NCMI evaluated with linear probing (\texttt{LP}) and with centroids comparison (\texttt{CC}), respectively.

\subsection{Experiments on CIFAR-100}

\begin{table}
\centering
\caption{Top-1 validation accuracy on CIFAR-100 for models trained with NCMI and baseline methods, averaged over three random seeds. CC denotes greedy prediction via nearest-centroid comparison; LP denotes linear probing. The best and second-best results are shown in \textbf{bold} and \underline{underlined}, respectively.}
\label{Tab:CIFAR100Result}
\resizebox{\linewidth}{!}{\begin{tabular}{ccccc}
\toprule
\rowcolor{mygray} \multicolumn{5}{c}{CIFAR-100}\\ \midrule
\multicolumn{1}{c|}{Model}      & ResNet-18 & ResNet-34 & ResNet-50 & ResNet-101 \\ 
\multicolumn{1}{c|}{Feature Dim.}  & 512 & 512 & 2048 & 2048 \\ \hline
\multicolumn{1}{c|}{CE}       & 75.44     & 76.42     & 76.96     & 77.39      \\
\multicolumn{1}{c|}{LS}         & 75.92     & 76.77     & 77.06     & 77.37      \\
\multicolumn{1}{c|}{AntiClass} & 76.28     & 76.31     & 76.30     & 76.69      \\
\multicolumn{1}{c|}{Squentropy} & 75.71     & 76.62     & 77.15     & 77.74      \\
\multicolumn{1}{c|}{SquareLoss} & 75.10     & 76.62     & 77.15     & 77.74      \\
\multicolumn{1}{c|}{PolyLoss}   & 75.59     & 76.87     & 76.46     & 77.77      \\
\multicolumn{1}{c|}{SupCon}     & 73.00     & 74.53     & 74.88     & 75.77      \\
\multicolumn{1}{c|}{SupCon (large BS)}     & -     & -     & 77.04     & -     \\
\multicolumn{1}{c|}{Focal Loss} & 76.34     & 76.62     & 77.32     & 77.76      \\\hline
\multicolumn{1}{c|}{\texttt{NCMI-CC} (ours)}       & \underline{76.45}     &  \underline{76.97}     &  \underline{77.50}     &  \underline{77.81}      \\ 
\multicolumn{1}{c|}{\texttt{NCMI-LP} (ours)}       & \textbf{76.94}     & \textbf{77.54}    & \textbf{77.76}    & \textbf{78.23}      \\ \bottomrule
\end{tabular}}
\end{table}
\begin{figure}
    \centering
\includegraphics[width=\columnwidth]{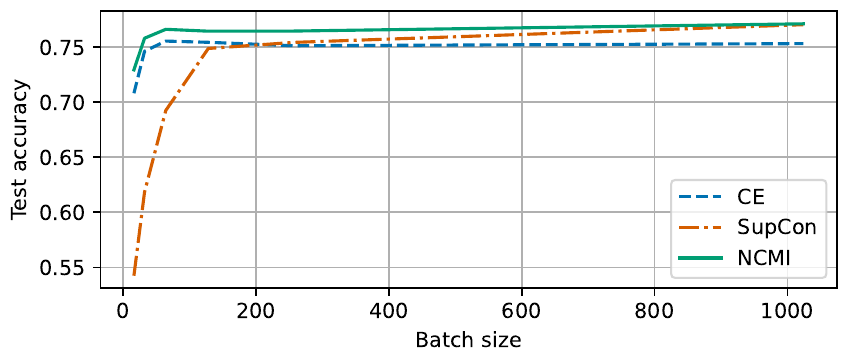}
     \caption{ResNet-50 test accuracy on CIFAR-100 as a function of batch size. We evaluate batch sizes \{16, 32, 64, 128, 256, 1024\}; NCMI consistently outperforms CE and SupCon across all settings.}
    \label{fig:acc_bs}
\end{figure}

The CIFAR-100 dataset contains 50-K training and 10-K test color images of resolution $32\times32$, which are labeled for 100 classes. 

To illustrate the effectiveness of NCMI, we evaluate NCMI on ResNet-18/34/50/101, whose penultimate feature dimensions are 512, 512, 2048, and 2048, respectively, and compare NCMI with respect to 8 benchmark methods namely, CE, LS \cite{szegedy2016rethinking}, AntiClass \cite{kat2025anticlasses}, Squentropy \cite{hui2023cut},  SquareLoss \cite{hui2021evaluation}, PolyLoss \cite{leng2022polyloss},  SupCon \cite{khosla2020supervised}  and Focal Loss \cite{lin2017focal}. 

For all surrogate losses, we use an SGD optimizer with a momentum of 0.9, a learning rate of 0.1, and a weight decay of 0.0005, along with a batch size of 64. We train the model for 200 epochs, and at epochs 60, 120, and 160, we reduce the current learning rate by a factor  of 10. Since SupCon relies on a large batch size to work, we report the results they reported in the paper and the reproduced results under the same setting.   

The results are reported in \Cref{Tab:CIFAR100Result}. As seen, NCMI yields consistent improvements across all backbones under both evaluation protocols. With linear probing, \texttt{NCMI-LP} achieves the best accuracy for every architecture, improving over CE by $+1.50\%$ (ResNet-18),  $+1.12\%$ (ResNet-34), $+0.80\%$ (ResNet-50), and $+0.84\%$ (ResNet-101). Moreover, the strongest gains occur for the smaller simplex dimension.


\subsection{Experiments on ImageNet}

\begin{table}
\centering
\caption{Top-1 and Top-5 validation accuracy on ImageNet for models trained with NCMI and baseline methods. CC denotes greedy prediction via nearest-centroid comparison; LP denotes linear probing. The best and second-best results are shown in \textbf{bold} and \underline{underlined}, respectively.}
\label{Tab:ImageNet}
\vspace{-0.2cm}
\begin{tabular}{ccccc}
\toprule
\rowcolor{mygray} \multicolumn{5}{c}{ImageNet}\\ \midrule
\multicolumn{1}{c|}{\multirow{2}{*}{Method}} & \multicolumn{2}{c|}{ResNet-50}     & \multicolumn{2}{c}{ResNet-101} \\
\multicolumn{1}{c|}{}                        & Top-1 & \multicolumn{1}{c|}{Top-5} & Top-1         & Top-5         \\ \hline
\multicolumn{1}{c|}{CE}                      & 76.24 & \multicolumn{1}{c|}{92.42} & 78.42         & 95.35\\
\multicolumn{1}{c|}{LS}                      & 78.37 & \multicolumn{1}{c|}{94.83} & 79.10         & 96.46         \\
\multicolumn{1}{c|}{Focal Loss}              & 78.11 & \multicolumn{1}{c|}{94.64} & 79.75         &   94.66     \\
\multicolumn{1}{c|}{SupCon}                  &  63.78 & \multicolumn{1}{c|}{86.60} &   67.43        &  90.24       \\ 
\multicolumn{1}{c|}{SupCon (large BS)}                  & 78.70 & \multicolumn{1}{c|}{94.30} & 79.33         &  94.52         \\ \hline
\multicolumn{1}{c|}{\texttt{NCMI-CC} (ours)}                    & \textbf{79.01} & \multicolumn{1}{c|}{\underline{95.34}} &   \textbf{79.97}       &    \textbf{96.64}   \\
\multicolumn{1}{c|}{\texttt{NCMI-LP} (ours)}                    & \underline{78.92} & \multicolumn{1}{c|}{\textbf{96.23}} &  \underline{79.83}       &    \underline{96.55}  \\ \bottomrule
\end{tabular}
\end{table}

ImageNet \cite{5206848} is a large-scale image recognition dataset that contains around 1.2M training samples and 50K validation images. We have conducted experiments on two models from the ResNet family, namely ResNet-50 and ResNet-101, and evaluated NCMI's performance against CE, LS, Focal Loss, and SupCon. Similar with the CIFAR-100 setting, we use an SGD optimizer with momentum of 0.9 learning rate of 0.5, weight decay of 5e-5 and batch size of 1024, we train the models with 1000 epochs, with cosine annealing learning rate decay, For all the methods, we train the model using the image resolution of $224\times 224$, while at evaluation, we apply a resolution of $ 280\times 280$. For SupCon, we report the results under the same setting as all other methods and those presented in their paper. 

The results are summarized in \Cref{Tab:ImageNet}. With ResNet-50, NCMI improves top-1 accuracy by 2.77\% over CE.

\subsection{Experiments on Whole Slide Image Dataset}

\begin{table}
\centering
\caption{Test F1 score and AUC on CAMELYON-17 and BRACS dataset for models trained with NCMI and baseline methods. LP denotes linear probing.}
\label{Tab:WSIresults}
\begin{tabular}{l|ll|ll}
\toprule
\rowcolor{mygray} Dataset  & \multicolumn{2}{l|}{CAMELYON-17} & \multicolumn{2}{l}{BRACS} \\ \midrule
Method   & F1 score $\uparrow$        & AUC  $\uparrow$ & F1 score  $\uparrow$ & AUC  $\uparrow$ \\ \hline
\rowcolor{lightgray}ABMIL    & 0.522            & 0.853         & 0.680         & 0.866     \\
\rowcolor{lightgray}\texttt{+NCMI-LP} & 0.567            & 0.892         & 0.701         & 0.872     \\
TransMIL & 0.554            & 0.792         & 0.631         & 0.841     \\
+\texttt{NCMI-LP} & 0.582            & 0.853         & 0.662         & 0.878     \\
\rowcolor{lightgray}AEM      & 0.647            & 0.887         & 0.742         & 0.905     \\
\rowcolor{lightgray}\texttt{+NCMI-LP} & 0.663            & 0.907         & 0.779         & 0.918     \\
ASMIL    & 0.689            & 0.898         & 0.781         & 0.914     \\
+\texttt{NCMI-LP} & 0.710            & 0.914         & 0.824         & 0.936     \\ \bottomrule
\end{tabular}
\end{table}

To assess NCMI beyond natural image datasets, we evaluate on two whole-slide image (WSI) classification benchmarks: CAMELYON-17 \cite{8447230} and BRACS \cite{brancati2022bracs}. CAMELYON-17 comprises 1000 WSIs from five medical centers, providing a diverse and clinically representative cohort. Of these, 500 slides are publicly available with slide-level labels, while the remaining 500 are held out for challenge evaluations. The multi-institutional composition introduces substantial variation in staining and scanning, making CAMELYON-17 a strong test bed for generalization. BRACS is a large-scale WSI dataset curated for breast cancer subtype classification, comprising 547 WSIs collected from several institutions and annotated by expert pathologists into clinically relevant categories: benign tumors, atypical tumors, and malignant tumors. We follow the official split of each dataset into training, validation, and test sets.

We compare four state-of-the-art multiple instance learning methods, namely, ASMIL \cite{ASMIL}, TransMIL \cite{shao2021transmil}, ABMIL \cite{ITW:2018} and AEM \cite{zhang2025AEM}. For each method, we remove the classification head and train with the NCMI surrogate loss. Then, we apply linear probing to all methods. Because both datasets are class-imbalanced, we use macro-averaged AUC and macro-averaged F1 as the primary metrics. Results in \Cref{Tab:WSIresults} show that replacing cross-entropy with NCMI consistently improves both F1 and AUC on CAMELYON-17 and BRACS. 
\begin{table}[t]
\centering
\caption{Per-epoch wall-clock time and peak graph memory for ResNet-50 and ResNet-101 on ImageNet.}
\label{tab:efficiency}
\resizebox{\linewidth}{!}{\begin{tabular}{ccccc}
\toprule
\rowcolor{mygray} \multicolumn{5}{c}{ImageNet}\\ \midrule
\multicolumn{1}{c|}{}            & \multicolumn{2}{c}{ResNet-50} & \multicolumn{2}{c}{ResNet-101} \\
\multicolumn{1}{c|}{}            & Time $\downarrow$         & Memory  $\downarrow$        & Time  $\downarrow$         & Memory  $\downarrow$ \\ \hline
\multicolumn{1}{c|}{CE}          & 6 mins 39 s & 102.29 Gb & 10 mins 33 s  &  142.67 Gb          \\
\multicolumn{1}{c|}{Focal Loss}  & 6 mins 42 s & 104.63 Gb & 10 mins 35 s &  143.42 Gb       \\
\multicolumn{1}{c|}{SupCon}      & 9 mins 52 s  & 180.32 Gb &   16 mins 03 s  & 241.17 Gb   \\
\multicolumn{1}{c|}{\texttt{NCMI} (ours)} & 6 mins 44 s  & 107.89 Gb & 10 mins 49 s & 148.55 Gb \\ \bottomrule
\end{tabular}}
\end{table}
\subsection{Training Cost \& Training Stability}
Compared with SupCon, NCMI uses less GPU memory and trains faster because it requires only a single forward pass and a simple objective. We quantify these efficiency gains on ImageNet in \Cref{tab:efficiency}, reporting per-epoch wall-clock time and peak GPU memory usage. For fairness, all experiments were conducted on a server with two \textit{AMD EPYC 7763 CPUs} and eight \textit{NVIDIA A5000 GPUs}, using the same optimizer and data pre-processing, with batch size of 1024, As seen, compare with SupCon, NCMI only take $59.83\%$ of the graphic memory, on par with the CE and Focal Loss, while largely outperform all the baselines in terms of classification accuracy.

NCMI also converges reliably with small batch size. \Cref{fig:acc_bs} shows how batch size affects the validation accuracy of CE, SupCon, and NCMI. We train ResNet-50 on CIFAR-100 with batch sizes \{16, 32, 64, 128, 256, 1024\}. Across all batch sizes, NCMI exhibits robust convergence and consistently surpasses CE.  Furthermore, we observe that SupCon relies heavily on large batches; reducing the batch size results in a significant decline in accuracy.
\begin{table}[t]
\centering
\caption{Component-wise ablation of NCMI on CIFAR-100. We evaluate the contribution of the NSF and the centering operation.}
\label{Tab:ablation}
\begin{tabular}{ccccc}
\toprule
\rowcolor{mygray} \multicolumn{5}{c}{CIFAR-100} \\ \midrule
\multicolumn{1}{c|}{NSF}       & \ding{51}      & \ding{51}    & \ding{55}    & \ding{55}    \\
\multicolumn{1}{c|}{Centering} & \ding{51}     & \ding{55}    & \ding{51}    & \ding{55}    \\ \hline
\multicolumn{1}{c|}{Accuracy}  & 76.45 & 74.9 & 1.72 & 5.64 \\ \bottomrule
\end{tabular}
\end{table}

\section{Ablation Study}
To understand the design choice, in this section, we evaluate the effects of the NSF and feature centering on the CIFAR-100 dataset by enabling and disabling them in all possible combinations, as shown in \Cref{Tab:ablation}. Removing either component degrades performance, with the NSF having the larger impact. Replacing the NSF with a softmax head causes the model to fail to converge to a non-trivial solution.

To better understand how these components influence training, we further assess the impact of NSF and the centering operation by visualizing the resulting class clusters and their corresponding centers. Due to space constraints, these results are provided in the appendix\footnote{Extended version (with appendix): https://arxiv.org/pdf/2601.02543.}.
\section{Conclusion}
In this paper, we present a new surrogate loss for DNN-based classifiers, called normalized conditional mutual information (NCMI). We further propose a novel alternating learning algorithm to minimize the NCMI loss to train a DNN-based classifier. Extensive experiment results over natural images and WSI datasets consistently show that DNN-based classifiers trained with NCMI outperform those trained using other CE-based or heuristic loss functions.

Open questions include: (1) how to extend the CMI and $\Gamma$ with multiple centroids per class to further improve the learning process, (2) how to develop a robust version of NCMI to improve adversarial robustness, and (3) how to extend NCMI to the natural language process under auto-regression pretraining. We leave these problems for future work.




\bibliographystyle{IEEEtran}
\bibliography{refs} 
\include{appendix}

\end{document}

%% file: appendix.tex
\definecolor{cmtgray}{gray}{0.55}
\definecolor{mygray}{gray}{.8}
\definecolor{lightgray}{gray}{0.93}
\definecolor{Mygray}{gray}{0.35}
\algrenewcommand\textproc{\textsc} 
\algrenewcommand\algorithmicrequire{\textsc{Input:}}
\algrenewcommand\algorithmicensure{\textsc{Output:}}
\algrenewcommand{\algorithmiccomment}[1]{\hfill\textcolor{cmtgray}{\#~#1}}
\interdisplaylinepenalty=2500 

\hyphenation{op-tical net-works semi-conduc-tor}







\appendices

\setcounter{section}{0} 
\section{Proof of \Cref{theorem1}}
We present the proof of \Cref{theorem1} in this section.

\begin{align}
    &\frac{\frac{1}{|\mathcal{D}|}\sum_{y} \sum_{\boldsymbol{x}\in \mathcal{D}^y} D(\boldsymbol{p_x}\|\boldsymbol{s}^y)}{\frac{1}{|\mathcal{D}|^2}\sum_{\boldsymbol{z}\in \mathcal{D}} \sum_{\boldsymbol{x}\in \mathcal{D}} \mathds{1}_{\{\boldsymbol{c_x}\neq \boldsymbol{c_z}\}} D(\boldsymbol{s^{\boldsymbol{c_x}}}\|\boldsymbol{p_z})} \nonumber\\ 
    =&\frac{\frac{1}{|\mathcal{D}|}\sum_{y} \sum_{\boldsymbol{x}\in \mathcal{D}^y} \Big[\sum_{i=1}^n \boldsymbol{p_x}[i] \ln \frac{\boldsymbol{p_x}[i]}{\boldsymbol{s}^y[i]} \Big]}{\frac{1}{|\mathcal{D}|^2}\sum_{\boldsymbol{z}\in \mathcal{D}} \sum_{\boldsymbol{x}\in \mathcal{D}}\mathds{1}_{\{\boldsymbol{c_x}\neq \boldsymbol{c_z}\}}\Big[\sum_{i=1}^n \boldsymbol{s}^{\boldsymbol{c_x}}[i] \ln \frac{\boldsymbol{s}^{\boldsymbol{c_x}}[i]}{\boldsymbol{p_z}[i]} \Big]}\\
    =& \min_{\boldsymbol{q}^v, v\in [C]} \frac{\frac{1}{|\mathcal{D}|}\sum_{y} \sum_{\boldsymbol{x}\in \mathcal{D}^y} \Big[\sum_{i=1}^n \boldsymbol{p_x}[i] \ln \frac{\boldsymbol{p_x}[i]}{\boldsymbol{q}^y[i]} \Big]}{\frac{1}{|\mathcal{D}|^2}\sum_{\boldsymbol{z}\in \mathcal{D}} \sum_{\boldsymbol{x}\in \mathcal{D}}\mathds{1}_{\{\boldsymbol{c_x}\neq \boldsymbol{c_z}\}}
    \Big[\sum_{i=1}^n \boldsymbol{s}^{\boldsymbol{c_x}}[i] \ln \frac{\boldsymbol{q}^{\boldsymbol{c_x}}[i]}{\boldsymbol{p_z}[i]} \Big]} \\
    =& \min_{\boldsymbol{q}^v, v\in [C]} \frac{\frac{1}{|\mathcal{D}|}\sum_{y} \sum_{\boldsymbol{x}\in \mathcal{D}^y} D(\boldsymbol{p_x}\|\boldsymbol{q}^y)}{\frac{1}{|\mathcal{D}|^2}\sum_{\boldsymbol{z}\in \mathcal{D}} \sum_{\boldsymbol{x}\in \mathcal{D}}\mathds{1}_{\{\boldsymbol{c_x}\neq \boldsymbol{c_z}\}}\Big[H(\boldsymbol{p_x}, \boldsymbol{p_z}) - H(\boldsymbol{p_x}, \boldsymbol{q^{c_z}})\Big]}
\end{align}

\setcounter{section}{1} 
\section{Extended Ablations: Feature Centering and Normalized Sigmoid (NSF)}
\begin{figure*}
    \centering
\includegraphics[width=\textwidth]{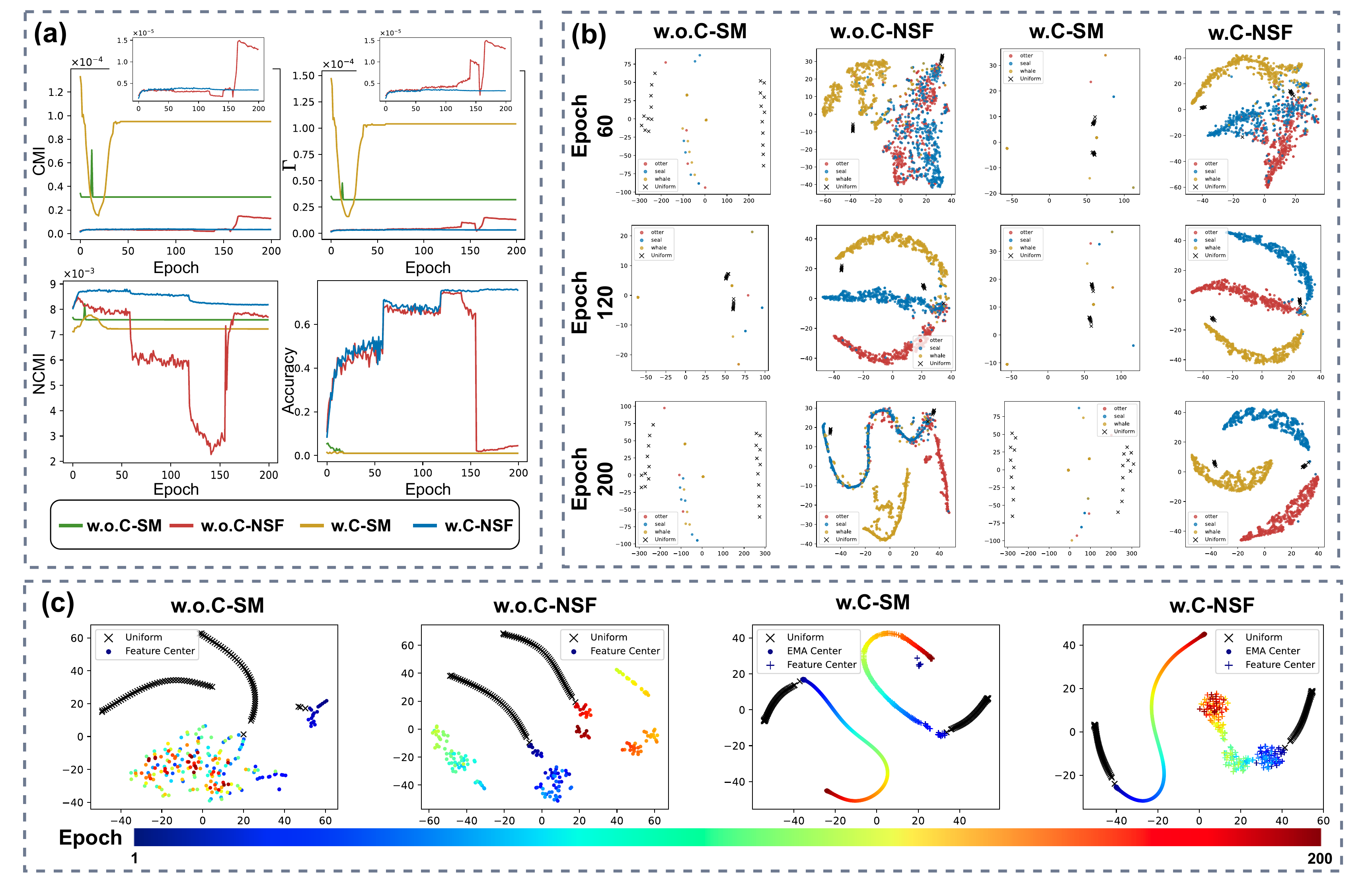}
\caption{Ablation of feature centering and the normalized sigmoid (NSF). We ablate each component by enabling or disabling it:  \textbf{w.o.C}/\textbf{w.C} denote without/with centering, and \textbf{SM}/\textbf{NSF} denote applying softmax/normalized sigmoid function. (a) Training curves of CMI, $\Gamma$, NCMI, and top-1 accuracy for ResNet-18 on CIFAR-100. (b) t-SNE of features from three randomly selected classes at epochs 60, 120, and 200; black crosses mark constant-valued vectors (all entries equal), which map via $\sigma$ to the uniform distribution. (c) t-SNE trajectories of feature centers and their EMA updates across training under all settings.}
    \label{fig:cluster_training}
\end{figure*}

Figure~\ref{fig:cluster_training} summarizes the effect of centering and NSF. Panel (a) shows the evolution of CMI, $\Gamma$, NCMI, and accuracy during training. Panel (b) visualizes feature clusters at epochs $\{60,120,200\}$ under each setting; the black crosses indicate constant-valued vectors that correspond, after $\sigma$, to the uniform distribution. Enabling centering pulls class clusters toward these reference points, thereby mitigating drift toward biased outputs. Panel (c) plots the trajectories of feature centers (and EMA-updated centers, when applicable). With centering, the centers remain stably concentrated around the constant-valued directions; without centering, they drift and collapse.

Softmax (SM) tends to produce degenerate manifolds in the t-SNE space—indicating that a few logits dominate the feature—whereas NSF suppresses overlarge entries and yields better-balanced probabilities, stabilizing optimization. 

Together, NSF and centering prevent single-mode collapse and avoid output distributions dominated by a few entries, leading to a more stable and reliable training process.

